\PassOptionsToPackage{table}{xcolor}
\documentclass[sigconf]{acmart}

\usepackage{multirow}
\usepackage{makecell}
\usepackage{graphicx}
\usepackage{subcaption}

\definecolor{modelblue}{RGB}{230,240,255} 

\AtBeginDocument{%
  }

\copyrightyear{2026}
\acmYear{2026}
\setcopyright{cc}
\setcctype{by}
\acmConference[ICMR '26]{International Conference on Multimedia Retrieval}{June 16--19, 2026}{Amsterdam, Netherlands}
\acmBooktitle{International Conference on Multimedia Retrieval (ICMR '26), June 16--19, 2026, Amsterdam, Netherlands}
\acmDOI{10.1145/3805622.3810577}
\acmISBN{979-8-4007-2617-0/2026/06}

\begin{document}

\title{RMPL: Relation-aware Multi-task Progressive Learning with Stage-wise Training for Multimedia Event Extraction}

\author{Yongkang Jin}
\email{im.jinyongkang@gmail.com}
\affiliation{%
  \institution{School of Computer Science and Technology, Soochow University}
  \city{Suzhou}
  \country{China}
}

\author{Jianwen Luo}
\email{jwluo.ai@gmail.com}
\affiliation{%
  \institution{School of Computer Science and Technology, Soochow University}
  \city{Suzhou}
  \country{China}
}

\author{Jingjing Wang}
\email{djingwang@suda.edu.cn}
\affiliation{%
  \institution{School of Computer Science and Technology, Soochow University}
  \city{Suzhou}
  \country{China}
}

\author{Jianmin Yao}
\email{jyao@suda.edu.cn}
\affiliation{%
  \institution{School of Computer Science and Technology, Soochow University}
  \city{Suzhou}
  \country{China}
}

\author{Yu Hong}
\email{tianxianer@gmail.com}
\authornote{Corresponding author.}
\affiliation{%
  \institution{School of Computer Science and Technology, Soochow University}
  \city{Suzhou}
  \country{China}
}


\begin{abstract}
Multimedia Event Extraction (MEE) aims to identify events and their arguments from documents that contain both text and images. It requires grounding event semantics across different modalities. Progress in MEE is limited by the lack of annotated training data. M2E2 is the only established benchmark, but it provides annotations only for evaluation. This makes direct supervised training impractical. Existing methods mainly rely on cross-modal alignment or inference-time prompting with Vision--Language Models (VLMs). These approaches do not explicitly learn structured event representations and often produce weak argument grounding in multimodal settings. To address these limitations, we propose RMPL, a Relation-aware Multi-task Progressive Learning framework for MEE under low-resource conditions. RMPL incorporates heterogeneous supervision from unimodal event extraction and multimedia relation extraction with stage-wise training. The model is first trained with a unified schema to learn shared event-centric representations across modalities. It is then fine-tuned for event mention identification and argument role extraction using mixed textual and visual data. Experiments on the M2E2 benchmark with multiple VLMs show consistent improvements across different modality settings.
\end{abstract}





\maketitle

\section{Introduction}

\begin{figure}[t]
    \centering
    \includegraphics[scale=0.68]{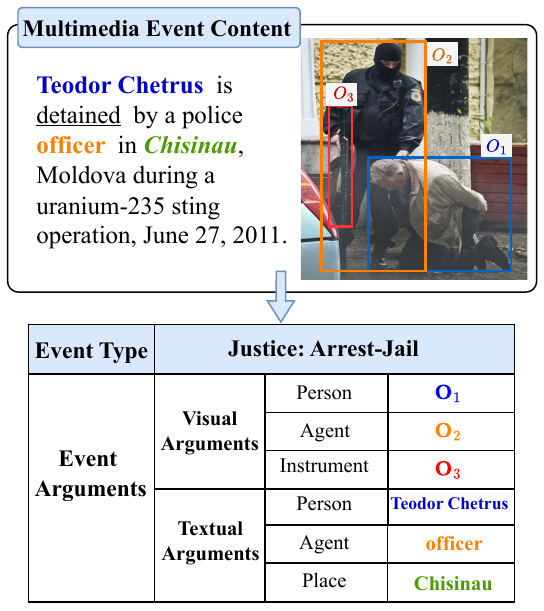}
    \caption{An example of MEE from M2E2~\cite{li-etal-2020-cross}. The \textit{Justice: Arrest-Jail} event is recognized from multimedia event content with paired text and image inputs, triggered by ``detained'', and event arguments are classified across modalities.}
    \label{fig:task-intro}
\end{figure}

Multimedia Event Extraction (MEE) is the task of identifying events and their argument structures from multimedia contents that contain both text and images~\cite{li-etal-2020-cross}. 
A single multimedia event content can involve multiple events, each associated with an event type and a set of arguments.
Compared with traditional text-based event extraction\cite{yang-etal-2024-scented, 10.1145/3652583.3658076, srivastava-etal-2025-instruction, liang-etal-2025-adaptive}, MEE requires jointly understanding textual descriptions and visual scenes, and assigning semantic roles to arguments grounded in different modalities. 
As shown in Figure~\ref{fig:task-intro}, the multimedia event content consists of a news sentence and an associated image describing an arrest scenario. 
From this multimedia event content, the model needs to identify the event type as \emph{Justice: Arrest-Jail}. 
Meanwhile, it should extract visual arguments from the image, including \emph{Person} as $O_{1}$, \emph{Agent} as $O_{2}$, and \emph{Instrument} as $O_{3}$, as well as textual arguments from the sentence, where \emph{Person} is ``Teodor Chetrus'', \emph{Agent} is ``officer'', and \emph{Place} is ``Chisinau''.

Most existing approaches to MEE treat this task as a cross-modal alignment or inference-based formulation. Early methods associate textual event mentions with visual evidence through weakly supervised or shared multimodal representations~\cite{li-etal-2020-cross}, and later work enhances such alignment using contrastive learning or event-aware pretraining~\cite{10.1145/3503161.3548132,9879585}. To alleviate the scarcity of annotated multimedia data, several studies introduce synthetic data generation, cross-modal multi-task learning, or unified inference frameworks~\cite{10.1145/3581783.3612526,Cao2025XMTL,seeberger-etal-2024-mmutf,10533232}. More recently, Vision--Language Models (VLMs) are applied to MEE~\cite{11210082,yu-etal-2025-multimedia,xing-etal-2025-benchmarking}, where most approaches rely on prompting and in-context learning at inference time. Although these methods show promising reasoning ability, they are not optimized through supervised fine-tuning, and therefore tend to focus on alignment rather than structured event modeling, which limits accurate grounding of argument roles and event structure.
 
A central challenge in MEE lies in the lack of annotated training data for MEE. 
To date, M2E2~\cite{li-etal-2020-cross} remains the only established benchmark for MEE, and it provides annotations exclusively for evaluation, without a training split, due to the high cost and complexity of multimodal event annotation. 
This setting prevents direct supervised training on the target task and forces models to rely on indirect supervision from heterogeneous external data sources. 
However, supervision from text-only or image-only event extraction datasets is inherently incomplete, as it captures event semantics within a single modality and does not explicitly model how event types and argument roles are jointly grounded across text and images. 
As a result, simply transferring unimodal event knowledge to MEE often leads to inconsistent role assignments or weak grounding of event arguments in multimodal contexts. 
Meanwhile, recent approaches based on VLMs mainly operate through inference-time prompting, which offers limited capability for systematically learning structured, event-centric representations under supervision. 
These limitations highlight the need for a training paradigm that can effectively and systematically leverage heterogeneous supervision while explicitly guiding models toward structured event understanding across modalities in this low resource setting.

To address this challenge, we propose \textbf{RMPL}, a \textbf{R}elation-aware \textbf{M}ulti-task \textbf{P}rogressive \textbf{L}earning framework with stage-wise training for MEE. 
Instead of relying on annotated multimedia event data, RMPL leverages high-quality unimodal event extraction datasets together with structured multimodal relation supervision. 
The framework follows a coarse-to-fine training paradigm. 
In the first stage, the model is warmed up under a unified schema with heterogeneous supervision from textual event extraction, visual event extraction, and multimedia relation extraction, enabling it to learn shared event-centric representations across modalities. 
In the second stage, the model is further specialized through task-specific supervised fine-tuning for event mention identification and argument role extraction using mixed textual and visual data. 
By explicitly modeling relational structure during warm-up and progressively specializing the model toward event-focused objectives, RMPL enables effective learning for MEE under realistic low-resource settings.

Our contributions are summarized as follows:
\begin{itemize}
    \item We identify a key bottleneck in current MEE research. M2E2 remains the only established benchmark and it provides no training split, which makes direct supervised training infeasible and motivates the need for a learning paradigm that can exploit heterogeneous external supervision.
    \item We propose \textbf{RMPL}, a relation-aware multi-task progressive learning framework with stage-wise training. The framework performs unified schema warm-up with textual event extraction, visual event extraction, and multimedia relation extraction, and then specializes the model through task-specific supervised fine-tuning for event mention identification and argument role extraction.
    \item We conduct experiments on M2E2 across multiple VLM backbones under a consistent evaluation protocol. The results show that RMPL achieves improvements across text-only, image-only and multimedia settings, and further comparisons and ablation studies validate the effectiveness of stage-wise specialization and supervision mixing strategies.
\end{itemize}

\section{Related Work}

\subsection{Multimedia Event Extraction}
MEE aims to identify events and their arguments from documents that jointly contain textual and visual information. This task was first systematically formulated by Li et al.~\cite{li-etal-2020-cross}, who established the M2E2 benchmark and provided an initial formalization of event extraction across text and images. Early approaches mainly relied on cross-modal alignment to transfer event semantics across modalities. Representative methods include FLAT~\cite{li-etal-2020-cross} and WASE~\cite{li-etal-2020-cross}, which learn weakly supervised structured embeddings to associate textual event mentions with visual evidence. 
Subsequent studies further improved cross-modal correspondence through more explicit alignment mechanisms, such as unified contrastive learning in UniCL~\cite{10.1145/3503161.3548132} and event-aware pretraining in CLIP-EVENT~\cite{9879585}. To address the scarcity of annotated multimedia event data, later work explored several directions, including synthetic data generation and curriculum training in CAMEL~\cite{10.1145/3581783.3612526}, cross-modal multi-task learning in X-MTL~\cite{Cao2025XMTL}, unified template formulation in MMUTF~\cite{seeberger-etal-2024-mmutf}, text-centric extraction enhanced by visual object representations in UMIE~\cite{Sun2024UMIE}, and coarse-to-fine inference in MGIM~\cite{10533232}.
Despite steady progress, these approaches mainly focus on cross-modal semantic alignment, data adaptation, or task unification, and do not explicitly model internal event structure among arguments or progressively specialize models toward event-centric objectives, which ultimately limits effective grounding of visual evidence to high-level event roles in MEE.

More recently, advances in VLMs have motivated a new line of research on applying foundation models to MEE.
Most existing studies primarily explore adaptation strategies at inference-time or with limited parameter updates, aiming to exploit the strong cross-modal reasoning capabilities of large models under constrained supervision.
For example, SSGPF~\cite{11210082} adopts a step-wise prompting strategy guided by event schemas with limited parameter updates, while KE-MME~\cite{yu-etal-2025-multimedia} applies knowledge editing to VLMs without additional parameter optimization.
In parallel, Xing et al.~\cite{xing-etal-2025-benchmarking} show that although VLMs demonstrate strong reasoning capability on M2E2, they still face clear limitations in semantic precision, argument grounding, and consistency.
Obviously, existing approaches built on VLMs mainly rely on prompting, in-context learning or limited adaptation and do not fully leverage supervised fine-tuning to explicitly enhance structured, event-centric representations.

\subsection{Multi-task Learning}

Multi-task learning has been widely studied as an effective strategy for leveraging related tasks to improve generalization and representation learning\cite{lu-etal-2022-unified, zhang-etal-2025-exploring, li2025mruiemultiperspectivereasoningreinforcement, fu-etal-2025-training}.
In multimedia understanding, this paradigm has been primarily explored by jointly optimizing structurally related information extraction tasks across multiple modalities.
For example, multimedia relation extraction has been studied under multi-task settings, where modeling structured relations across text and images provides complementary semantic cues for cross-modal reasoning~\cite{Zheng2021MultimodalRE,zhang-etal-2025-exploring}.
UMIE~\cite{Sun2024UMIE} further unifies multiple multimodal information extraction tasks, including named entity recognition, relation extraction, and event extraction, within a single framework by adaptively integrating visual object information into text-centric representations.
In contrast, X-MTL~\cite{Cao2025XMTL} adopts a cross-modal multi-task learning strategy that jointly optimizes event extraction objectives on textual and visual data sources, where the multi-task formulation mainly arises from modality and dataset heterogeneity rather than from jointly modeling distinct structural tasks.
In textual event extraction, relation-aware and multi-task learning have also been shown to benefit event extraction by explicitly modeling dependencies among event arguments and promoting shared representations across related subtasks, such as REAR~\cite{luo-etal-2025-rear}, which leverages relation semantics for event argument disambiguation, or InstructUIE~\cite{Wang2023InstructUIEMI}, which unifies multiple information extraction tasks under a shared schema.
However, existing approaches either focus on unimodal settings or adopt multi-task learning across modalities, without explicitly modeling event structures and their associated components within a unified framework for MEE.

\section{Task Formulation}
\label{sec:task-formulation}
MEE is the task of identifying events and their argument structures from documents containing both text and images. A multimedia document is represented as $\mathcal{D}=\langle \mathcal{S}, \mathcal{M}\rangle$, where $\mathcal{S}=\{s_1,\dots,s_{|\mathcal{S}|}\}$ denotes a set of textual sentences and $\mathcal{M}=\{m_1,\dots,m_{|\mathcal{M}|}\}$ denotes a set of associated images. 
Under this formulation, MEE consists of two core subtasks: event mention identification and argument role extraction. 
In addition, we introduce multimedia relation extraction as an auxiliary task to provide structured relational supervision.

\paragraph{\textbf{Event Mention Identification.}}
Event Mention Identification requires predicting the event type $y_e$ of an event mention from a predefined event type set $\mathcal{Y}_E$, given a multimedia document $\mathcal{D}$.
An event mention may be supported by textual evidence $s\in\mathcal{S}$, visual evidence $m\in\mathcal{M}$, or both. A multimedia event is supported by both textual and visual evidence $(s,m)$, while a text-only event or an image-only event is supported by a single modality $(s,\varnothing)$ or $(\varnothing,m)$.

\paragraph{\textbf{Argument Role Extraction.}}
Argument Role Extraction is defined as the task of identifying event arguments and assigning each argument a semantic role with respect to a given event type. Given a multimedia document $\mathcal{D}$ and an event type $y_e$, the task outputs a set of argument--role pairs describing the participants involved in the event.
In text, arguments correspond to entity mentions in sentences $s\in\mathcal{S}$, while in images, arguments correspond to objects grounded by bounding boxes in images $m\in\mathcal{M}$. An event argument may be text-only $(s,\varnothing)$, image-only $(\varnothing,m)$, or multimedia $(s,m)$.

\paragraph{\textbf{Multimedia Relation Extraction.}}
Multimedia Relation Extraction aims to identify semantic relations between pairs of entities or objects in multimedia documents containing text and images. Given a document $\mathcal{D}$ consisting of a sentence and an associated image, the task predicts relation triples of the form $\langle \text{head}, \text{relation}, \text{tail} \rangle$, where the head and tail may appear in text, image, or both modalities. 
In our framework, multimedia relation extraction is incorporated as an auxiliary objective that provides explicit relational supervision to support structured event-centric representation learning.


\begin{figure*}[t]
    \centering
    \includegraphics[width=\textwidth]{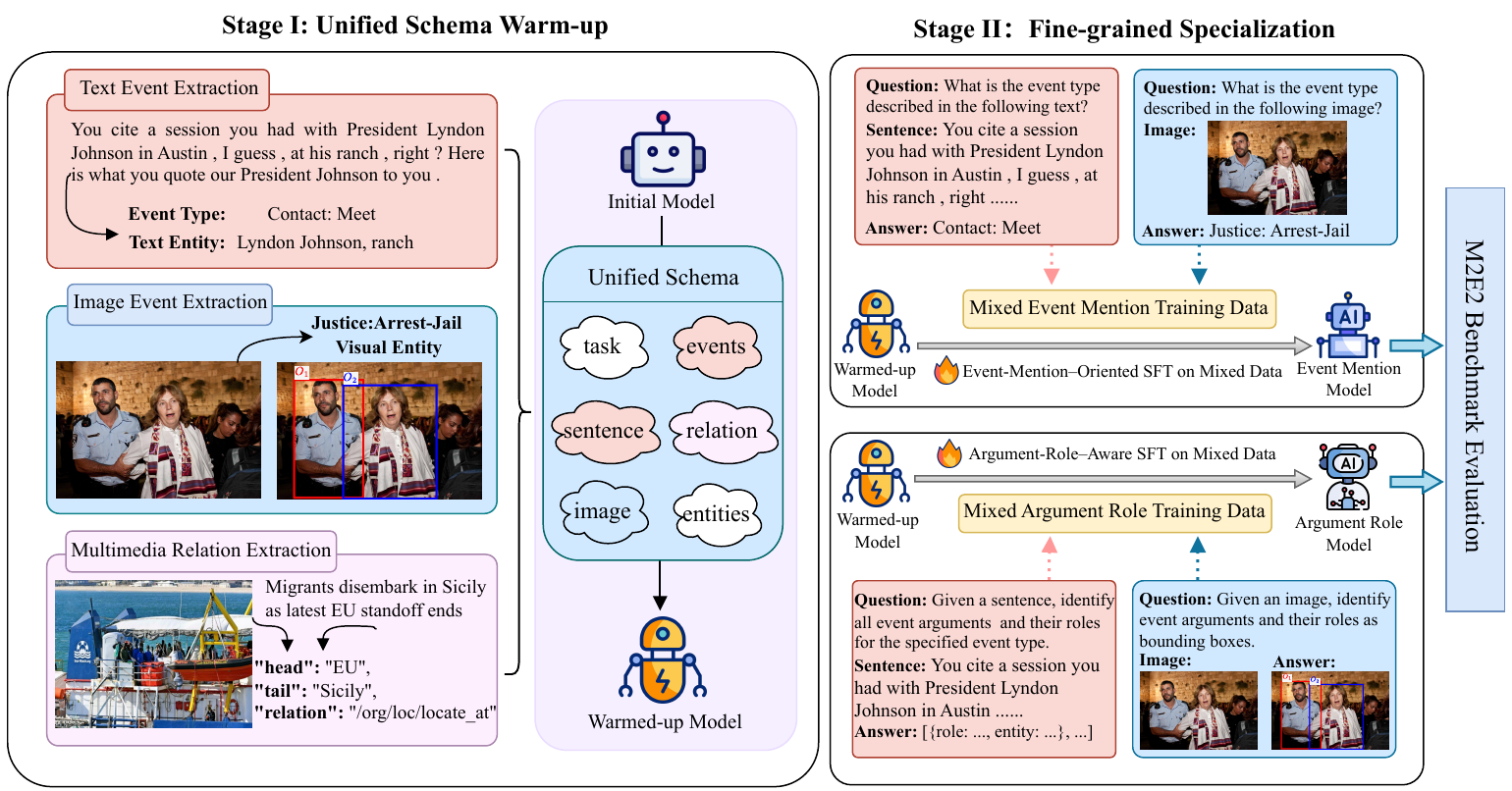}
    \caption{
        Overview of RMPL.
        Stage I conducts unified schema warm-up training with heterogeneous event-centric supervision to learn general event representations.
        Stage II further performs task-specific supervised training for event mention identification and argument role extraction across modalities, and finally evaluates the resulting models on the M2E2 benchmark.
    }
    \label{fig:overview}
\end{figure*}

\section{Approach}
In this section, we introduce RMPL, a relation-aware multi-task progressive learning framework with stage-wise training for MEE.

\subsection{Overview}
\label{sec:overview}
Figure~\ref{fig:overview} illustrates the architecture design of RMPL for MEE.
Specifically, RMPL adopts a coarse-to-fine, stage-wise learning paradigm and leverages heterogeneous event-centric supervision from multiple sources, including textual event extraction, visual event extraction, and multimedia relation extraction.
By unifying these signals within a shared schema, RMPL progressively improves event-centric understanding and cross-modal grounding.

The left panel of Figure~\ref{fig:overview} illustrates Stage I, referred to as Unified Schema Warm-up. 
Starting from an initial VLM backbone, we perform warm-up training under a unified schema that explicitly aligns core semantic elements involved in event understanding, including task, events, sentence, relation, image, and entities. 
In this stage, the model is jointly trained with heterogeneous supervision from three sources: textual event extraction, which provides event type and argument annotations from textual descriptions; visual event extraction, which associates visual entities with event semantics in images; and multimedia relation extraction, which supplies structured relational signals that link entities across text and image modalities.
Through this unified schema warm-up, the model learns shared event-centric representations that capture structured semantic regularities across modalities, resulting in a general event-aware information extraction model. 
Correspondingly, the right panel of Figure~\ref{fig:overview} depicts Stage II, termed Fine-grained Specialization. 
Building upon the warmed-up model obtained in Stage I, we construct mixed task-specific training data for two downstream tasks: event mention identification and argument role extraction. 
For each task, supervision from both text and image modalities is combined to form mixed training sets, allowing the model to adapt to modality-specific characteristics while retaining the shared representations learned during the warm-up stage.
The model is then fine-tuned separately for each task using task-oriented supervised fine-tuning on the corresponding mixed textual and visual data, and the resulting models are evaluated on the M2E2~\cite{li-etal-2020-cross} benchmark to assess the effectiveness and generalization in realistic MEE scenarios.

\subsection{Stage I: Unified Schema Warm-up}
The objective of Stage~I is to initialize the VLM backbone with structure-aware, event-centric representations under heterogeneous supervision. 
Training under a unified schema enables the model to capture shared semantic regularities across modalities, serving as a foundation for subsequent task-specific specialization.

Formally, following the task formulation in Section~\ref{sec:task-formulation}, 
each training instance corresponds to a multimedia document $\mathcal{D}$ with associated textual and visual content. 
Stage~I adopts a conditional generation formulation based on a predefined schema to generate structured event or relation representations from $\mathcal{D}$.
To unify heterogeneous supervision under the same generation interface, we introduce a schema control variable $\tau \in \{t,i,r\}$ that specifies the schema family to be generated, corresponding to textual event extraction, visual event extraction, and multimedia relation extraction, respectively. 
Under a fixed serialization protocol, the target schema is linearized into a token sequence $Y=(y_1,\dots,y_{|Y|})$, and the active slots as well as the output layout of $Y$ are determined by $\tau$. 
The model then performs autoregressive decoding to generate $Y$, factorizing the conditional likelihood as follows:
\begin{equation}
p_\theta\!\left(Y \mid \mathcal{D}, \tau\right)
=
\prod_{n=1}^{|Y|}
p_\theta\!\left(y_n \mid y_{<n}, \mathcal{D}, \tau\right),
\quad
\tau \in \{t,i,r\}.
\end{equation}
where $Y$ is the linearized schema sequence, $y_{<n}$ denotes the previously generated prefix, $p_\theta(\cdot)$ denotes the model conditional distribution, and $\tau$ controls the schema specification.

For event extraction supervision, we consider two schema types with $\tau \in \{t,i\}$, corresponding to textual and visual event extraction.
Under each source, the target output is a linearized event schema sequence $Y$ defined by the corresponding schema type. 
Given a training instance $(\mathcal{D}, Y)$ from $\mathcal{S}_{\tau}$, the model is trained to maximize the conditional log-likelihood of $Y$ given $\mathcal{D}$.
We define the event extraction loss as: 
\begin{equation}
\mathcal{L}_{\tau}
=
\mathbb{E}_{(\mathcal{D},Y)\sim \mathcal{S}_{\tau}}
\Big[
-\log p_{\theta}\!\left(Y \mid \mathcal{D}, \tau\right)
\Big],
\quad
\tau \in \{t, i\}.
\end{equation}
where $\mathcal{S}_{\tau}$ denotes the training data distribution 
associated with $\tau$.
For relation supervision corresponding to $\tau=r$, each training instance is associated with a set of annotated relation triples $\{\langle h,\rho,u\rangle\}$ derived from the multimedia document $\mathcal{D}$, where $h$ and $u$ denote the head and tail entities, and $\rho$ denotes the semantic relation between them. 
The model is trained to maximize the conditional log-likelihood of these relation triples given the input document and the relation schema, leading to the relation extraction loss:
\begin{equation}
\resizebox{0.92\hsize}{!}{$
\mathcal{L}_{r}
=
\mathbb{E}_{(\mathcal{D},\mathcal{R})\sim \mathcal{S}_{r}}
\Big[
-\log p_{\theta}\!\left(
\mathcal{R}
\mid
\mathcal{D}, \tau=r
\right)
\Big],
\quad
\mathcal{R}=\{\langle h_j,\rho_j,u_j\rangle\}_{j=1}^{|\mathcal{R}|},
$}
\end{equation}
where $\mathcal{S}_{r}$ denotes the training data distribution associated with $r$.

The overall Stage~I training objective is a weighted combination of the heterogeneous supervision signals from the three sources:
\begin{equation}
\mathcal{L}_{\mathrm{I}}
=
\alpha\,\mathcal{L}_{t}
+
\beta\,\mathcal{L}_{i}
+
\lambda\,\mathcal{L}_{r},
\end{equation}
where $\alpha$, $\beta$, and $\lambda$ control the relative mixing proportions of the three supervision sources during Stage~I training, and the impact of different mixing proportions is further analyzed in Section~\ref{sec:ablation:mixing}. 

\subsection{Stage II: Fine-grained Specialization}

The objective of Stage II is to specialize the warmed-up model from Stage I for MEE by performing task-specific fine-tuning on event mention identification and argument role extraction. 
Starting from the warmed-up model in Stage I, Stage II focuses on refining task-specific decision boundaries using mixed textual and visual supervision, while multimedia relation extraction is no longer involved at this stage due to the task-specific focus of Stage II.

Formally, in Stage~II, we consider two task-specific objectives corresponding to event mention identification and argument role extraction. 
For event mention identification, each training instance is associated with a textual or visual input and an event type label. 
Conditioned on the modality-specific input representation, the model predicts a categorical distribution over event types. 
The learning objective is defined as the expected negative log-likelihood of the ground-truth event type under textual and visual supervision:
\begin{equation}
\mathcal{L}_{\text{EM}}
=
\mathbb{E}_{(X^{\tau}, y_e)}
\Big[
-\log p_\theta(y_e \mid X^{\tau})
\Big], \quad \tau \in \{t, i\},
\end{equation}
where $X^{t}=\mathcal{S}$ and $X^{i}=\mathcal{M}$ denote the textual and visual inputs, respectively, and $y_e \in \mathcal{Y}_E$ is the event type annotation.

For argument role extraction, the model is further optimized to assign semantic roles to event arguments conditioned on the event type and the modality-specific input. 
Each training instance consists of an input representation, an event type label, and a set of annotated argument role pairs. 
The argument role extraction objective is defined as the expected negative log-likelihood of the ground-truth role assignments over all arguments:
\begin{equation}
\resizebox{0.92\hsize}{!}{$
\mathcal{L}_{\text{AR}}
=
\mathbb{E}_{(X^{\tau}, y_e, A)}
\;
\mathbb{E}_{(a_k, y_k)\sim A}
\Big[
-\log p_\theta(y_k \mid a_k, X^{\tau}, y_e)
\Big],
~~ \tau \in \{t, i\} ,
$}
\end{equation}
where $X^{t}=\mathcal{S}$ and $X^{i}=\mathcal{M}$ denote the textual and visual inputs, $A$ denotes the annotated argument entity mention-role pairs in the input text, and $a_k$ denotes an argument entity mention, with $y_k \in \mathcal{Y}_R$ its ground-truth role label.

Collectively, Stage~II optimizes the model for MEE by jointly considering event mention identification and argument role extraction in a coordinated manner. 
Accordingly, the overall training objective of Stage~II is defined as the sum of the two task-specific losses:
\begin{equation}
\mathcal{L}_{\mathrm{II}}
=
\mathcal{L}_{\text{EM}}
+
\mathcal{L}_{\text{AR}}.
\end{equation}

By jointly optimizing the two task-specific objectives under a shared and unified multimodal backbone, Stage~II further sharpens task-specific decision boundaries for event mention identification and argument role extraction in a coordinated and targeted manner.  
This fine-grained task specialization further enhances prediction accuracy while preserving the general event-centric representations learned during Stage~I, thereby enabling reliable MEE in realistic settings with limited task-specific annotations.

\section{Experimentation}

In this section, we evaluate the effectiveness of RMPL. 
We first describe the datasets and experimental settings, and then present the main results on the M2E2 benchmark. 
Finally, we compare RMPL with previous methods and conduct ablation studies to analyze both individual components and supervision mixing proportions.

\subsection{Experimental Setup}

\begin{table}[t]
\caption{Statistics of datasets used in our experiments.}
\centering
\small
\setlength{\tabcolsep}{2.8pt} 
\begin{tabular}{lccccc}
\toprule
\textbf{Dataset}
& \textbf{\#Sents}
& \textbf{\#Images}
& \textbf{\#Types}
& \textbf{\#Roles} 
& \textbf{\#Relations}
\\
\midrule
M2E2~\cite{li-etal-2020-cross}
& 6{,}167
& 1{,}014
& 8
& 15 
& -- \\

ACE 2005~\cite{doddington-etal-2004-automatic}
& 15{,}789
& --
& 33
& 36 
& -- \\

imSitu~\cite{7780966}
& --
& 126{,}102
& 504
& 1{,}788 
& -- \\

MNRE~\cite{10.1145/3474085.3476968}
& 9{,}201
& 9{,}201
& --
& -- 
& 23 \\

\bottomrule
\end{tabular}
\label{tab:dataset_statistics}
\end{table}


\paragraph{\textbf{Datasets.}}
We conduct our experiments on the M2E2~\cite{li-etal-2020-cross} benchmark and leverage ACE 2005~\cite{doddington-etal-2004-automatic}, SWiG~\cite{10.1007/978-3-030-58548-8_19}, and MNRE~\cite{10.1145/3474085.3476968} as external supervision sources during training, where SWiG is further augmented with object grounding annotations from imSitu~\cite{7780966}.
The dataset statistics are summarized in Table~\ref{tab:dataset_statistics}.
Specifically, M2E2 defines 8 event types and 15 argument roles, containing 1{,}297 textual event mentions and 391 visual event mentions, of which 192 textual and 203 visual mentions are aligned to form 309 multimedia events.
Since M2E2 does not provide annotated training data, we follow previous work~\cite{li-etal-2020-cross,10.1145/3503161.3548132,10.1145/3581783.3612526,10533232,Cao2025XMTL} and rely on external supervision from ACE 2005 and SWiG for model training, further incorporating MNRE as an additional source of relational supervision.
Following Li et al.~\cite{li-etal-2020-cross}, we map both textual event types and visual activity semantics to the same 8 event types defined in M2E2.

\paragraph{\textbf{Experimental Settings.}}
To ensure a fair comparison, all experiments are conducted under the same experimental and evaluation protocol of Yu et al.~\cite{yu-etal-2025-multimedia}. 
In addition, consistent with prior work~\cite{yu-etal-2025-multimedia,xing-etal-2025-benchmarking}, we adopt Qwen2-VL-7B~\cite{wang2024qwen2vlenhancingvisionlanguagemodels} as the backbone model in the main experiments.
Under this protocol, all images in the M2E2 benchmark are resized to a fixed resolution of $512 \times 512$ at evaluation time, and models are evaluated in \textit{text-only}, \textit{image-only}, and \textit{multimedia} settings, reporting Precision (P), Recall (R), and F1-score (F1) for \textit{event mention identification} and \textit{argument role extraction}, with \textit{macro-F1} and \textit{micro-F1} used respectively.
For visual arguments, a predicted bounding box is considered correct if its IoU with the ground-truth exceeds 0.5. 
During training, Stage~I is conducted for 400 steps with a learning rate of 1e-5, while Stage~II is trained for 800 steps with a learning rate of 3e-5, using a batch size of 16 for both stages. 
In addition to Qwen2-VL-7B, we further evaluate our framework on Qwen3-VL-8B~\cite{bai2025qwen3vltechnicalreport} and InternVL3\_5-8B~\cite{wang2025internvl35advancingopensourcemultimodal}\footnote{\url{https://huggingface.co/OpenGVLab/InternVL3_5-8B-HF}} to examine generalization across different backbone architectures, and all experiments are conducted on two NVIDIA RTX 4090 GPUs.

\begin{table*}[!t]
\centering
\small
\setlength{\tabcolsep}{3.5pt}
\renewcommand{\arraystretch}{1.3}
\caption{Performance comparison on the M2E2 benchmark under text-only, image-only, and multimedia settings. 
Results are reported for different backbone models under the Baseline and Warm-up variants, as well as the full RMPL model, evaluated using P, R, and F1, with macro-F1 and micro-F1 used for event mention and argument role evaluation.}
\label{tab:m2e2_main_llm}
\resizebox{\textwidth}{!}{%
\begin{tabular*}{1.00\textwidth}{@{\extracolsep{\fill}}p{0.08\textwidth}|ccc|ccc|ccc|ccc|ccc|ccc@{}}
\toprule
\multicolumn{1}{c|}{\smash{\raisebox{-5.5ex}{\textbf{Method}}}}
& \multicolumn{9}{c|}{\textbf{Event Mention}}
& \multicolumn{9}{c}{\textbf{Argument Role}} \\
\cmidrule(lr){2-10}\cmidrule(lr){11-19}
& \multicolumn{3}{c|}{\textbf{text-only}}
& \multicolumn{3}{c|}{\textbf{image-only}}
& \multicolumn{3}{c|}{\textbf{multimedia}}
& \multicolumn{3}{c|}{\textbf{text-only}}
& \multicolumn{3}{c|}{\textbf{image-only}}
& \multicolumn{3}{c}{\textbf{multimedia}} \\
\cmidrule(lr){2-4}\cmidrule(lr){5-7}\cmidrule(lr){8-10}
\cmidrule(lr){11-13}\cmidrule(lr){14-16}\cmidrule(lr){17-19}
& P & R & F1 & P & R & F1 & P & R & F1
& P & R & F1 & P & R & F1 & P & R & F1 \\

\midrule
\rowcolor{modelblue}
\multicolumn{19}{c}{\textbf{Qwen2-VL-7B-Instruct}} \\
\multicolumn{1}{c|}{Baseline}
& 73.54 & 62.68 & 63.63 & 76.28 & 70.73 & 67.36 & 83.75 & 81.81 & 78.04
& 18.34 & 51.24 & 27.01 & 13.48 & 29.93 & 18.59 & 17.15 & 32.00 & 22.33 \\
\multicolumn{1}{c|}{Warm-up}
& 75.38 & 64.40 & 66.70 & 78.65 & 73.62 & 69.76 & 88.09 & 73.64 & 76.90 
& 18.72 & 52.14 & 27.55 & 20.42 & 44.41 & 27.98 & 26.32 & 38.50 & 31.26 \\
\multicolumn{1}{c|}{RMPL}
& \textbf{85.63} & \textbf{71.27} & \textbf{77.07} & \textbf{80.14} & \textbf{77.88} & \textbf{75.80} & \textbf{89.84} & \textbf{89.54} & \textbf{88.21} 
& \textbf{48.85} & \textbf{59.13} & \textbf{53.50} & \textbf{55.22} & \textbf{48.68} & \textbf{51.75} & \textbf{42.93} & \textbf{47.72} & \textbf{45.20} \\

\midrule
\rowcolor{modelblue}
\multicolumn{19}{c}{\textbf{InternVL3\_5-8B}} \\
\multicolumn{1}{c|}{Baseline}
& 73.73 & 66.51 & 66.81 & 69.00 & 83.68 & 74.06 & \textbf{89.11} & \textbf{88.56} & 86.63
& 26.22 & 64.92 & 37.35 & 24.07 & 42.43 & 30.71 & 21.74 & 33.62 & 26.41 \\
\multicolumn{1}{c|}{Warm-up}
& 74.80 & 67.38 & 67.82 & 68.70 & \textbf{85.76} & 74.76 & 82.61 & 88.32 & 82.44
& 26.10 & \textbf{65.58} & 37.34 & 26.97 & \textbf{48.36} & 34.63 & 22.35 & 34.60 & 27.16 \\
\multicolumn{1}{c|}{RMPL}
& \textbf{83.68} & \textbf{69.52} & \textbf{75.28} & \textbf{80.16} & 79.60 & \textbf{79.05} & 88.44 & \textbf{88.56} & \textbf{87.05}
& \textbf{48.89} & 58.35 & \textbf{53.20} & \textbf{40.00} & 45.39 & \textbf{42.53} & \textbf{42.36} & \textbf{49.89} & \textbf{45.82} \\

\midrule
\rowcolor{modelblue}
\multicolumn{19}{c}{\textbf{Qwen3-VL-8B-Instruct}} \\
\multicolumn{1}{c|}{Baseline}
& 76.49 & 65.71 & 66.91 & 75.39 & 76.36 & 73.69 & 88.83 & 82.20 & 80.46
& 25.25 & 63.53 & 36.13 & 47.29 & 66.12 & 55.14 & 40.16 & \textbf{47.83} & 43.66 \\
\multicolumn{1}{c|}{Warm-up}
& 76.73 & 65.16 & 66.50 & 72.24 & \textbf{86.08} & 77.43 & 88.94 & 82.11 & 80.29
& 26.17 & \textbf{64.50} & 37.23 & 45.61 & \textbf{68.42} & 54.74 & 41.06 & 46.85 & 43.77 \\
\multicolumn{1}{c|}{RMPL}
& \textbf{82.73} & \textbf{70.96} & \textbf{75.90} & \textbf{83.56} & 81.01 & \textbf{81.73} & \textbf{93.36} & \textbf{92.51} & \textbf{92.44}
& \textbf{50.61} & 57.56 & \textbf{53.86} & \textbf{66.41} & 57.24 & \textbf{61.48} & \textbf{49.58} & 44.58 & \textbf{46.94} \\

\bottomrule
\end{tabular*}
}
\end{table*}

\subsection{Main Results}
\label{sec:main_results}

\paragraph{\textbf{Baselines.}}
Table~\ref{tab:m2e2_main_llm} summarizes the performance of three representative VLM backbones evaluated under different training stages within a unified evaluation protocol, including
(i) Baseline, which conducts prompt-only inference on M2E2 without any supervised fine-tuning;
(ii) Warm-up, which employs unified schema multi-task learning with heterogeneous supervision from unimodal event extraction and multimedia relation extraction;
(iii) RMPL, which incorporates explicit task-specific supervised fine-tuning for event mention identification and argument role extraction, starting from the warmed-up model without relying on annotated MEE data.

\paragraph{\textbf{Overall Experimental Results.}}
Table~\ref{tab:m2e2_main_llm} summarizes the performance of different VLM backbones under the Baseline and Warm-up configurations, as well as the full RMPL model.
In general, RMPL consistently improves performance across backbone architectures and evaluation settings, indicating reliable and generalizable gains.
By organizing model training into an initial warm-up stage followed by task-specific specialization, our approach enables more effective learning without relying on annotated MEE training data.
The Warm-up stage introduces unified event-centric supervision with relation-aware signals, regularizing representation learning across heterogeneous modalities, as further supported by the component-wise ablation results in Section~\ref{sec:ablation:component}. 
Although the performance gains achieved at this stage are generally moderate, the warm-up process stabilizes training and promotes the learning of more structured and transferable representations that are essential for subsequent task-specific optimization. 
In a small number of cases, slight performance fluctuations are observed for multimedia event mention identification, which likely result from the emphasis of the Warm-up stage on schema-level consistency and cross-task generalization rather than direct optimization of task-specific evaluation objectives. 
In summary, the Warm-up stage offers a reliable initialization that consistently improves performance across multimodal tasks and settings when combined with RMPL.

Building upon the Warm-up initialization, RMPL delivers clear absolute performance gains over both the Baseline and the Warm-up variant across most subtasks. 
For event mention identification, RMPL achieves consistent F1 improvements under text-only, image-only, and multimedia settings for all three VLM backbones, with particularly strong gains observed in image-only and multimedia scenarios. For instance, absolute F1 improvements in multimedia event mention identification reach 10.17\% on Qwen2-VL-7B and 11.98\% on Qwen3-VL-8B.
For argument role extraction, the improvements are even more pronounced, with absolute F1 gains of up to 22.87\%, 19.41\%, and 3.28\% on Qwen2-VL-7B, InternVL3\_5-8B, and Qwen3-VL-8B, respectively, where the smaller margin on Qwen3-VL-8B reflects its inherently strong multimodal modeling capacity.
In summary, these results demonstrate that RMPL benefits from its stage-wise training strategy, in which the Warm-up stage establishes event-centric representations and the subsequent specialization stage enables task-specific optimization for MEE.

\begin{table*}[!htbp]
\centering
\small
\setlength{\tabcolsep}{4.2pt}
\renewcommand{\arraystretch}{1.3}
\caption{
    Performance comparison on M2E2 for event mention identification and argument role extraction under text-only, image-only, and multimedia settings.
    For RMPL, Ma, Mb, and Mc correspond to Qwen2-VL-7B, InternVL3\_5-8B, and Qwen3-VL-8B, respectively.
    Bold numbers indicate the best result in each column, while \underline{underlined} numbers denote the second-best.
}

\label{tab:m2e2_main}
\begin{tabular*}{0.98\textwidth}{@{\extracolsep{\fill}}
>{\centering\arraybackslash}p{0.13\textwidth}
|ccc|ccc|ccc|ccc|ccc|ccc@{}}
\toprule
\smash{\raisebox{-4.5ex}{\textbf{Method}}}
& \multicolumn{9}{c|}{\textbf{Event Mention}}
& \multicolumn{9}{c}{\textbf{Argument Role}} \\
\cmidrule(lr){2-10}\cmidrule(lr){11-19}

& \multicolumn{3}{c|}{\textbf{text-only}}
& \multicolumn{3}{c|}{\textbf{image-only}}
& \multicolumn{3}{c|}{\textbf{multimedia}}
& \multicolumn{3}{c|}{\textbf{text-only}}
& \multicolumn{3}{c|}{\textbf{image-only}}
& \multicolumn{3}{c}{\textbf{multimedia}} \\

\cmidrule(lr){2-4}\cmidrule(lr){5-7}\cmidrule(lr){8-10}
\cmidrule(lr){11-13}\cmidrule(lr){14-16}\cmidrule(lr){17-19}

& P & R & F1 & P & R & F1 & P & R & F1
& P & R & F1 & P & R & F1 & P & R & F1 \\
\midrule

\textbf{FLAT}~\cite{li-etal-2020-cross}
& 34.2 & 63.2 & 44.4
& 27.1 & 57.3 & 36.7
& 33.9 & 59.8 & 42.2
& 20.1 & 27.1 & 23.1
& 4.3 & 8.9 & 5.8
& 12.9 & 17.6 & 14.9 \\

\textbf{WASE}$_{\text{att}}$~\cite{li-etal-2020-cross}
& 37.6 & 66.8 & 48.1
& 32.3 & 63.4 & 42.8
& 38.2 & 67.1 & 49.1
& 27.5 & 33.2 & 30.1
& 9.7 & 11.1 & 10.3
& 18.6 & 21.6 & 19.9 \\

\textbf{WASE}$_{\text{obj}}$~\cite{li-etal-2020-cross}
& 42.8 & 61.9 & 50.6
& 43.1 & 59.2 & 49.9
& 43.0 & 62.1 & 50.8
& 23.5 & 30.3 & 26.4
& 14.5 & 10.1 & 11.9
& 19.5 & 18.9 & 19.2 \\

\textbf{CLIP-EVENT}~\cite{9879585}
& -- & -- & --
& 41.3 & 72.8 & 52.7
& -- & -- & --
& -- & -- & --
& 21.1 & 13.1 & 17.1
& -- & -- & -- \\

\textbf{UniCL}~\cite{10.1145/3503161.3548132}
& 49.1 & 59.2 & 53.7
& 54.6 & 60.9 & 57.6
& 44.1 & 67.7 & 53.4
& 27.8 & 34.3 & 30.7
& 16.9 & 13.8 & 15.2
& 24.3 & 22.6 & 23.4 \\

\textbf{CAMEL}~\cite{10.1145/3581783.3612526}
& 45.1 & \underline{71.8} & 55.4
& 52.1 & 66.8 & 58.5
& 55.6 & 59.5 & 57.5
& 24.8 & 41.8 & 31.1
& 21.4 & 28.4 & 24.4
& 31.4 & 35.1 & 33.2 \\

\textbf{UMIE}~\cite{Sun2024UMIE}
& -- & -- & --
& -- & -- & --
& -- & -- & 62.1
& -- & -- & --
& -- & -- & --
& -- & -- & 24.5 \\

\textbf{MGIM}~\cite{10533232}
& 50.1 & 66.5 & 55.8
& 55.7 & 64.4 & 58.5
& 46.3 & 69.6 & 55.6
& 28.2 & 34.7 & 31.2
& 24.1 & 14.1 & 17.8
& 25.2 & 21.7 & 24.6 \\

\textbf{MMUTF}~\cite{seeberger-etal-2024-mmutf}
& 48.5 & 65.0 & 55.5
& 55.1 & 59.1 & 57.0
& 47.9 & 63.4 & 54.6
& 33.6 & 44.2 & 38.2
& 23.6 & 18.8 & 20.9
& 39.9 & 20.8 & 27.4 \\

\textbf{X-MTL}~\cite{Cao2025XMTL}
& 49.7 & 65.7 & 56.6
& 73.1 & 70.3 & 71.7
& 78.3 & 57.3 & 66.2
& 34.6 & 37.6 & 36.0
& 33.2 & 31.3 & 32.2
& 40.3 & 42.6 & 41.4 \\

\textbf{FSMEE}~\cite{xing-etal-2025-benchmarking}
& 13.3 & 19.7 & 15.9
& 69.6 & 62.1 & 65.6
& 65.4 & 53.7 & 60.0
& \textbf{75.0} & 2.6 & 5.1
& 3.3 & 4.3 & 3.8
& 20.6 & 21.9 & 21.2 \\

\textbf{SSGPF}~\cite{11210082}
& -- & -- & --
& -- & -- & --
& 60.4 & 72.1 & 65.7
& -- & -- & --
& -- & -- & --
& 33.8 & 38.5 & 36.0 \\

\textbf{KE-MME}~\cite{yu-etal-2025-multimedia}
& 82.1 & \textbf{76.1} & \underline{76.0}
& 69.3 & 65.8 & 65.3
& \underline{89.8} & 87.7 & 87.4
& \underline{51.9} & 37.5 & 39.3
& 27.3 & 27.6 & 27.3
& \underline{48.1} & 43.9 & 44.3 \\

\midrule
\multirow{3}{*}{%
\begin{tabular}[c]{@{}@{\hspace{-2pt}}c@{\hspace{8pt}}|@{\hspace{12pt}}c@{}}
\renewcommand{\arraystretch}{1.35}%
\multirow{3}{*}{\textbf{RMPL}} & $\mathbf{M_a}$ \\
                              & $\mathbf{M_b}$ \\
                              & $\mathbf{M_c}$ \\
\end{tabular}%
}
& \textbf{85.6} & 71.3 & \textbf{77.1} & 80.1 & 77.9 & 75.8 & \underline{89.8} & \underline{89.5} & \underline{88.2}
& 48.9 & \textbf{59.1} & \underline{53.5} & \underline{55.2} & \underline{48.7} & \underline{51.8} & 42.9 & \underline{47.7} & 45.2  \\

& \underline{83.7} & 69.5 & 75.3
& \underline{80.2} & \underline{79.6} & \underline{79.1}
& 88.4 & 88.6 & 87.1
& 48.9 & \underline{58.4} & 53.2
& 40.0 & 45.4 & 42.5
& 42.4 & \textbf{49.9} & \underline{45.8} \\

& 82.7 & 71.0 & 75.9
& \textbf{83.6} & \textbf{81.0} & \textbf{81.7}
& \textbf{93.4} & \textbf{92.5} & \textbf{92.4}
& 50.6 & 57.6 & \textbf{53.9}
& \textbf{66.4} & \textbf{57.2} & \textbf{61.5}
& \textbf{49.6} & 44.6 & \textbf{46.9} \\

\bottomrule
\end{tabular*}
\end{table*}

\subsection{Comparison with Previous Work}
Table~\ref{tab:m2e2_main} presents a comparison between RMPL and prior methods on the M2E2 benchmark under text-only, image-only, and multimedia settings for both event mention identification and argument role extraction. 
These methods employ a diverse set of backbone architectures for MEE.
CLIP-EVENT~\cite{9879585} is built upon CLIP\cite{radford2021learningtransferablevisualmodels}, while UniCL~\cite{10.1145/3503161.3548132}, CAMEL~\cite{10.1145/3581783.3612526}, and X-MTL~\cite{Cao2025XMTL} adopt modular backbones that combine textual encoders such as BERT\cite{devlin-etal-2019-bert} with visual encoders including CLIP\cite{radford2021learningtransferablevisualmodels} or ViT\cite{DosovitskiyB0WZ21}, together with object detectors such as YOLOv8\cite{10533619}.
UMIE~\cite{Sun2024UMIE} adopts FLAN-T5\cite{chung2022scalinginstructionfinetunedlanguagemodels} as a unified text-centric backbone with visual features integrated into the language model, and MMUTF~\cite{seeberger-etal-2024-mmutf} further extends this design by incorporating T5\cite{JMLR:v21:20-074} with CLIP\cite{radford2021learningtransferablevisualmodels} and YOLOv8\cite{10533619} for multimodal fusion.
More recent approaches leverage VLM backbones, including SSGPF~\cite{11210082}, which is built on LLaVA-v1.5\cite{Liu_2024_CVPR}, and FSMEE~\cite{xing-etal-2025-benchmarking} and KE-MME~\cite{yu-etal-2025-multimedia}, both of which are based on Qwen2-VL-7B~\cite{wang2024qwen2vlenhancingvisionlanguagemodels} and evaluated under the same backbone in our experimental setting.
Given the substantial differences in backbone architectures and training setups, the comparison should be interpreted as a high-level reference rather than a strictly controlled benchmark, while still providing meaningful insights into performance trends in MEE.

Compared with prior approaches, RMPL achieves the best overall performance across different evaluation settings.
As shown in Table~\ref{tab:m2e2_main}, RMPL consistently outperforms existing methods on both event mention identification and argument role extraction, with particularly strong improvements under image-only and multimedia settings, where effective cross-modal grounding plays a critical role in accurate event understanding.
Notably, RMPL attains the highest F1 scores in the multimedia scenario, surpassing state-of-the-art methods such as KE-MME\cite{yu-etal-2025-multimedia} and X-MTL\cite{Cao2025XMTL}, indicating a more reliable and effective integration of textual and visual evidence for event understanding.
This advantage stems from RMPL, which introduces a warm-up stage with explicit relation modeling and applies supervised training to learn richer event-centric semantics beyond simple modality alignment or inference level adaptation.
Collectively, these results suggest that explicitly modeling structured semantics and using progressive training lead to a more generalizable approach for MEE under limited supervision.

\begin{table*}[t]
\centering
\setlength{\tabcolsep}{3.5pt}
\caption{
    \textbf{Ablation study of RMPL.}
    Results are reported for the Baseline, Warm-up variants with and without relation extraction supervision (RE), RMPL without the Warm-up stage, RMPL without relation extraction supervision, and the full RMPL model.
}
\label{tab:ablation_full}
\small
\setlength{\tabcolsep}{3.0pt}
\renewcommand{\arraystretch}{1.3}
\resizebox{\textwidth}{!}{%
\begin{tabular*}{1.00\textwidth}{@{\extracolsep{\fill}}p{0.12\textwidth}|ccc|ccc|ccc|ccc|ccc|ccc@{}}
\toprule
\multicolumn{1}{c|}{\smash{\raisebox{-5.5ex}{\textbf{Method}}}}
& \multicolumn{9}{c|}{\textbf{Event Mention}}
& \multicolumn{9}{c}{\textbf{Argument Role}} \\
\cmidrule(lr){2-10}\cmidrule(lr){11-19}
& \multicolumn{3}{c|}{\textbf{text-only}}
& \multicolumn{3}{c|}{\textbf{image-only}}
& \multicolumn{3}{c|}{\textbf{multimedia}}
& \multicolumn{3}{c|}{\textbf{text-only}}
& \multicolumn{3}{c|}{\textbf{image-only}}
& \multicolumn{3}{c}{\textbf{multimedia}} \\
\cmidrule(lr){2-4}\cmidrule(lr){5-7}\cmidrule(lr){8-10}
\cmidrule(lr){11-13}\cmidrule(lr){14-16}\cmidrule(lr){17-19}
& P & R & F1 & P & R & F1 & P & R & F1
& P & R & F1 & P & R & F1 & P & R & F1 \\
\midrule

Baseline
& 73.54 & 62.68 & 63.63 & 76.28 & 70.73 & 67.36 & 83.75 & 81.81 & 78.04
& 18.34 & 51.24 & 27.01 & 13.48 & 29.93 & 18.59 & 17.15 & 32.00 & 22.33 \\

Warm-up (w/o RE)
& 73.83 & 63.82 & 65.40 & 77.80 & 72.45 & 68.21 & 74.99 & 62.03 & 64.75
& 18.42 & 51.90 & 27.19 & 20.78 & 45.72 & 28.57 & 24.48 & 39.70 & 30.29 \\

Warm-up (w/ RE)
& 75.38 & 64.40 & 66.70 & 78.65 & 73.62 & 69.76 & 88.09 & 73.64 & 76.90
& 18.72 & 52.14 & 27.55 & 20.42 & 44.41 & 27.98 & 26.32 & 38.50 & 31.26 \\

\addlinespace
\midrule
\addlinespace

RMPL (w/o Warm)
& 83.94 & 69.08 & 74.83 & \textbf{81.85} & \textbf{78.67} & 74.35 & 88.53 & 88.28 & 86.49
& \textbf{49.85} & 58.71 & \textbf{53.92} & \textbf{55.56} & 46.05 & 50.36 & 42.44 & 44.47 & 43.43 \\

RMPL (w/o RE)
& \textbf{85.69} & 70.87 & 76.70 & 76.88 & 74.32 & 72.19 & 89.21 & 88.16 & 86.55
& 48.45 & 58.23 & 52.89 & 53.12 & 44.74 & 48.57 & 41.72 & 43.17 & 42.43 \\

RMPL (full)
& 85.63 & \textbf{71.27} & \textbf{77.07} & 80.14 & 77.88 & \textbf{75.80} & \textbf{89.84} & \textbf{89.54} & \textbf{88.21} 
& 48.85 & \textbf{59.13} & 53.50 & 55.22 & \textbf{48.68} & \textbf{51.75} & \textbf{42.93} & \textbf{47.72} & \textbf{45.20} \\

\bottomrule
\end{tabular*}
}
\end{table*}

\subsection{Extended Experiments}
To further analyze the effects of different components and training strategies in RMPL, 
we conduct a comprehensive extended experimental study on the M2E2 benchmark based on Qwen2-VL-7B.
The experiments are organized into two parts: 
(1) a component-wise analysis that evaluates the role of relation-aware supervision and stage-wise training strategies, and 
(2) an investigation of how different mixing proportions in Stage~I affect overall performance.

\begin{figure}[t]
    \centering
    \begin{subfigure}[t]{0.82\linewidth}
        \centering
        \includegraphics[width=\linewidth]{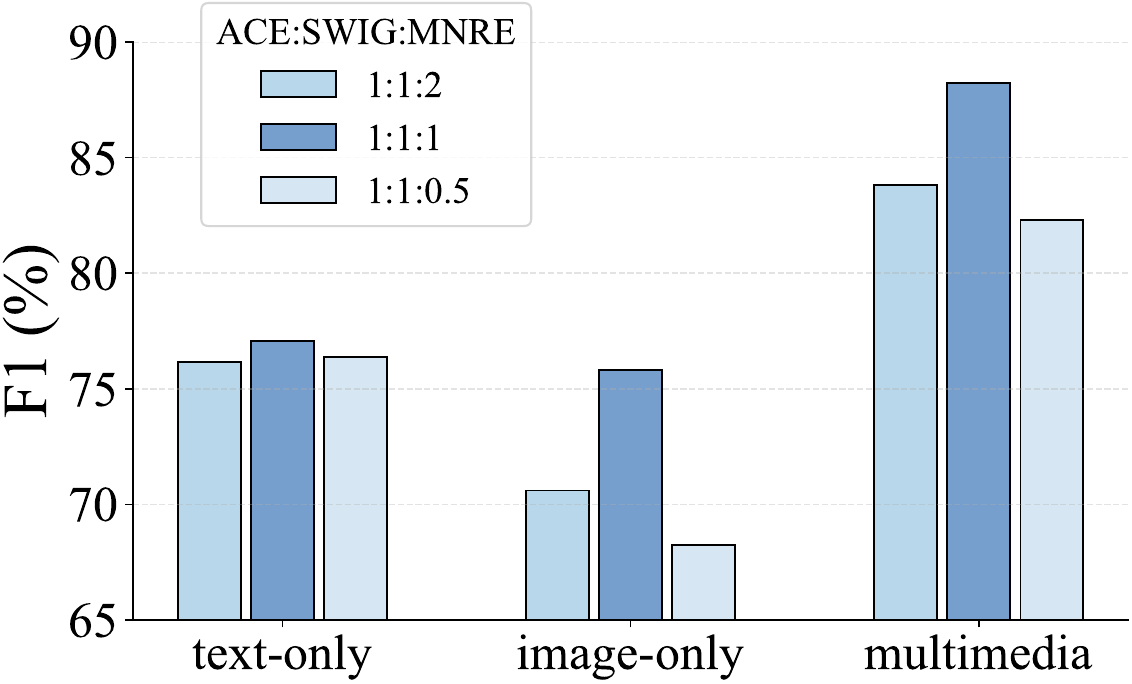}
        \caption{Event mention identification performance.}
        \label{fig:training_curve:em}
    \end{subfigure}
    \\
    \begin{subfigure}[t]{0.82\linewidth}
        \centering
        \includegraphics[width=\linewidth]{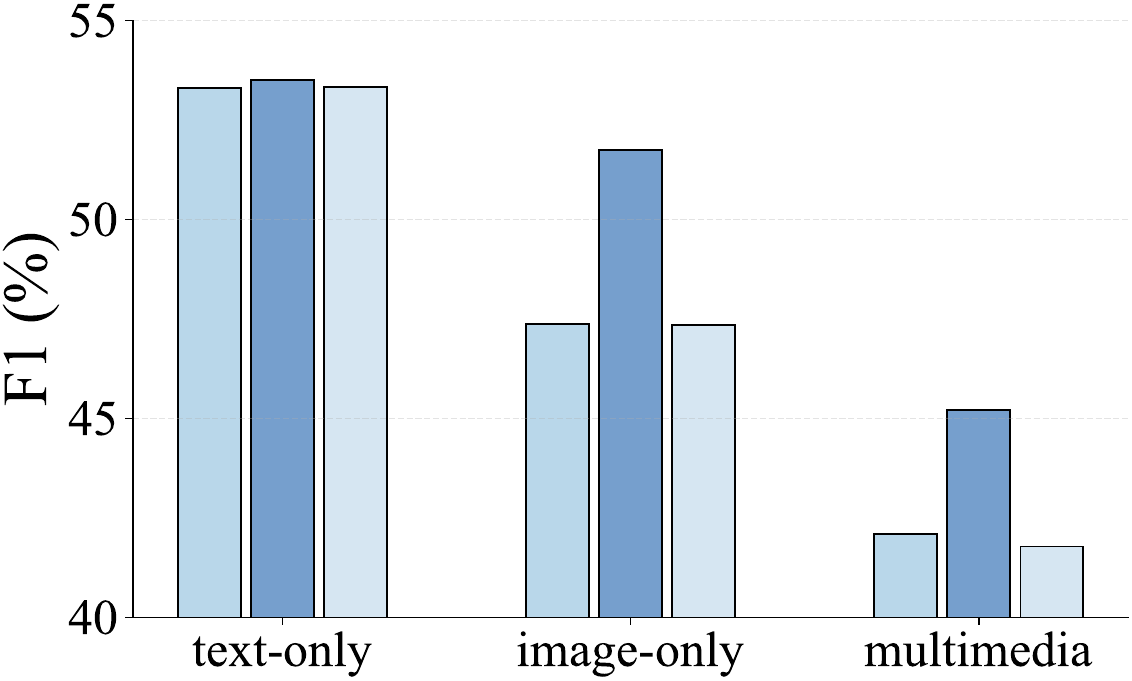}
        \caption{Argument role extraction performance.}
        \label{fig:training_curve:ar}
    \end{subfigure}

    \caption{Impact of different supervision mixing proportions.}
    \label{fig:training_curve}
\end{figure}

\subsubsection{Component-wise Ablation}
\label{sec:ablation:component}

Table~\ref{tab:ablation_full} reports the component-wise ablation results under text-only, image-only, and multimedia settings for both event mention identification and argument role extraction.
We evaluate a series of variants of RMPL by removing relation extraction supervision or different training stages to examine the contribution of each component.
The \textit{Baseline} denotes the original model without progressive training.
\textit{Warm-up (w/o RE)} and \textit{Warm-up (w/ RE)} correspond to the initialization stage of RMPL trained without and with relation extraction supervision, respectively.
\textit{RMPL (w/o Warm)} removes the Warm-up stage and directly performs task-specific supervised training.
\textit{RMPL (w/o RE)} applies task-specific supervised training on \textit{Warm-up (w/o RE)}.

Comparing \textit{Warm-up (w/o RE)} and \textit{Warm-up (w/ RE)} first demonstrates that incorporating relation extraction supervision during initialization yields consistent improvements, indicating that relation modeling enhances cross-modal alignment and supports the learning of more structured and semantically coherent event-centric representations. Extending this observation to the full framework, the comparison between \textit{RMPL (w/o RE)} and \textit{RMPL (full)} further shows that removing relation extraction supervision leads to additional performance degradation after task-specific training, underscoring its sustained contribution throughout the two-stage learning process. Finally, contrasting \textit{RMPL (w/o Warm)} with \textit{RMPL (full)} highlights the importance of the Warm-up stage itself, as eliminating progressive initialization weakens the overall effectiveness and stability of the model, particularly in argument role extraction.

\subsubsection{Impact of Mixing Proportions}
\label{sec:ablation:mixing}
In addition to the component-wise analysis, we further examine the impact of different supervision mixing proportions during the Warm-up stage.
Specifically, the sampling ratios of training instances from ACE, SWiG, and MNRE, corresponding to textual, visual, and relational supervision, are varied while keeping all other training settings fixed.
We consider three mixing strategies: a balanced setting (1:1:1), a relation-dominant setting (1:1:2), and a setting with increased unimodal event supervision (1:1:0.5).
Figure~\ref{fig:training_curve} illustrates the resulting performance trends under text-only, image-only, and multimedia evaluation settings.

We observe that the balanced 1:1:1 setting consistently yields the best performance across different evaluation settings and tasks.
We attribute this to the fact that such a ratio allows the model to learn strong unimodal event extraction capabilities from textual and visual data, while leveraging multimedia relation extraction to regulate and align representations across modalities.
Under this balanced supervision, the model captures modality-specific event semantics and mitigates modality bias through cross-modal relational signals during the Warm-up stage, resulting in more coherent event-centric representations.
In contrast, in the 1:1:0.5 setting, the reduced contribution of relational supervision weakens cross-modal interaction modeling, causing unimodal event recognition to dominate and limiting effective modality alignment, which affects fine-grained argument role extraction.
Conversely, in the 1:1:2 setting, excessive relational supervision biases the model toward abstract inter-entity patterns and diminishes sensitivity to modality-specific event grounding, resulting in degraded performance, especially in visual and multimedia scenarios.
These results suggest that balanced supervision is important for effective initialization of RMPL.

\section{Conclusion}

In this paper, we propose RMPL, a relation-aware multi-task progressive learning framework that leverages heterogeneous supervision and stage-wise training to learn structured, event-centric representations for MEE. By explicitly incorporating structured semantic signals during training, RMPL reduces reliance on annotated multimedia data and supports reliable learning under realistic low-resource conditions. Experiments on the M2E2 benchmark with multiple VLM backbones show consistent improvements across different modality settings, demonstrating strong generalization. 

In the future, we plan to extend RMPL to broader multimedia information extraction scenarios and incorporate a wider set of objectives into the unified training framework, enabling richer joint supervision, stronger cross-task knowledge sharing, and more reliable representation learning across modalities.

\begin{acks}
The research is supported by National Science Foundation of China  (62376182).
\end{acks}


\begin{thebibliography}{10}

\bibitem{li-etal-2020-cross}
Manling Li, Alireza Zareian, Qi~Zeng, Spencer Whitehead, Di~Lu, Heng Ji, and Shih-Fu Chang.
\newblock Cross-media structured common space for multimedia event extraction.
\newblock In Dan Jurafsky, Joyce Chai, Natalie Schluter, and Joel Tetreault, editors, {\em Proceedings of the 58th Annual Meeting of the Association for Computational Linguistics}, pages 2557--2568, Online, July 2020. Association for Computational Linguistics.

\bibitem{yang-etal-2024-scented}
Yu~Yang, Jinyu Guo, Kai Shuang, and Chenrui Mao.
\newblock Scented-{EAE}: Stage-customized entity type embedding for event argument extraction.
\newblock In Lun-Wei Ku, Andre Martins, and Vivek Srikumar, editors, {\em Findings of the Association for Computational Linguistics: ACL 2024}, pages 5222--5235, Bangkok, Thailand, August 2024. Association for Computational Linguistics.

\bibitem{10.1145/3652583.3658076}
Xintao Jiao, Jiansheng Chen, and Jiale Liu.
\newblock A graph convolution network with a pos-aware filter and context enhancement mechanism for event detection.
\newblock In {\em Proceedings of the 2024 International Conference on Multimedia Retrieval}, ICMR '24, page 285–292, New York, NY, USA, 2024. Association for Computing Machinery.

\bibitem{srivastava-etal-2025-instruction}
Saurabh Srivastava, Sweta Pati, and Ziyu Yao.
\newblock Instruction-tuning {LLM}s for event extraction with annotation guidelines.
\newblock In Wanxiang Che, Joyce Nabende, Ekaterina Shutova, and Mohammad~Taher Pilehvar, editors, {\em Findings of the Association for Computational Linguistics: ACL 2025}, pages 13055--13071, Vienna, Austria, July 2025. Association for Computational Linguistics.

\bibitem{liang-etal-2025-adaptive}
Sheng Liang, Hang Lv, Zhihao Wen, Yaxiong Wu, Yongyue Zhang, Hao Wang, and Yong Liu.
\newblock Adaptive schema-aware event extraction with retrieval-augmented generation.
\newblock In Christos Christodoulopoulos, Tanmoy Chakraborty, Carolyn Rose, and Violet Peng, editors, {\em Findings of the Association for Computational Linguistics: EMNLP 2025}, pages 7927--7946, Suzhou, China, November 2025. Association for Computational Linguistics.

\bibitem{10.1145/3503161.3548132}
Jian Liu, Yufeng Chen, and Jinan Xu.
\newblock Multimedia event extraction from news with a unified contrastive learning framework.
\newblock In {\em Proceedings of the 30th ACM International Conference on Multimedia}, MM '22, page 1945–1953, New York, NY, USA, 2022. Association for Computing Machinery.

\bibitem{9879585}
Manling Li, Ruochen Xu, Shuohang Wang, Luowei Zhou, Xudong Lin, Chenguang Zhu, Michael Zeng, Heng Ji, and Shih-Fu Chang.
\newblock Clip-event: Connecting text and images with event structures.
\newblock In {\em 2022 IEEE/CVF Conference on Computer Vision and Pattern Recognition (CVPR)}, pages 16399--16408, 2022.

\bibitem{10.1145/3581783.3612526}
Zilin Du, Yunxin Li, Xu~Guo, Yidan Sun, and Boyang Li.
\newblock Training multimedia event extraction with generated images and captions.
\newblock In {\em Proceedings of the 31st ACM International Conference on Multimedia}, MM '23, page 5504–5513, New York, NY, USA, 2023. Association for Computing Machinery.

\bibitem{Cao2025XMTL}
Jianwei Cao, Yanli Hu, Zhen Tan, and Xiang Zhao.
\newblock Cross-modal multi-task learning for multimedia event extraction.
\newblock In {\em Proceedings of the AAAI Conference on Artificial Intelligence}, volume~39, pages 11454--11462, 2025.

\bibitem{seeberger-etal-2024-mmutf}
Philipp Seeberger, Dominik Wagner, and Korbinian Riedhammer.
\newblock {MMUTF}: Multimodal multimedia event argument extraction with unified template filling.
\newblock In Yaser Al-Onaizan, Mohit Bansal, and Yun-Nung Chen, editors, {\em Findings of the Association for Computational Linguistics: EMNLP 2024}, pages 6539--6548, Miami, Florida, USA, November 2024. Association for Computational Linguistics.

\bibitem{10533232}
Yang Liu, Fang Liu, Licheng Jiao, Qianyue Bao, Long Sun, Shuo Li, Lingling Li, and Xu~Liu.
\newblock Multi-grained gradual inference model for multimedia event extraction.
\newblock {\em IEEE Transactions on Circuits and Systems for Video Technology}, 34(10):10507--10520, 2024.

\bibitem{11210082}
Xiang Yuan, Xinrong Chen, Haochen Li, Hang Yang, Guanyu Wang, Weiping Li, and Tong Mo.
\newblock Stepwise schema-guided prompting framework with parameter efficient instruction tuning for multimedia event extraction.
\newblock In {\em 2025 IEEE International Conference on Multimedia and Expo (ICME)}, pages 1--6, 2025.

\bibitem{yu-etal-2025-multimedia}
Jiaao Yu, Yijing Lin, Zhipeng Gao, Xuesong Qiu, and Lanlan Rui.
\newblock Multimedia event extraction with {LLM} knowledge editing.
\newblock In Christos Christodoulopoulos, Tanmoy Chakraborty, Carolyn Rose, and Violet Peng, editors, {\em Proceedings of the 2025 Conference on Empirical Methods in Natural Language Processing}, pages 4116--4124, Suzhou, China, November 2025. Association for Computational Linguistics.

\bibitem{xing-etal-2025-benchmarking}
Fuyu Xing, Zimu Wang, Wei Wang, and Haiyang Zhang.
\newblock Benchmarking and improving {LVLM}s on event extraction from multimedia documents.
\newblock In {\em Proceedings of the 18th International Natural Language Generation Conference}, pages 734--742, Hanoi, Vietnam, October 2025. Association for Computational Linguistics.

\bibitem{Sun2024UMIE}
Lin Sun, Kai Zhang, Qingyuan Li, and Renze Lou.
\newblock Umie: Unified multimodal information extraction with instruction tuning.
\newblock In {\em Proceedings of the AAAI Conference on Artificial Intelligence}, volume~38, pages 19062--19070, 2024.

\bibitem{lu-etal-2022-unified}
Yaojie Lu, Qing Liu, Dai Dai, Xinyan Xiao, Hongyu Lin, Xianpei Han, Le~Sun, and Hua Wu.
\newblock Unified structure generation for universal information extraction.
\newblock In Smaranda Muresan, Preslav Nakov, and Aline Villavicencio, editors, {\em Proceedings of the 60th Annual Meeting of the Association for Computational Linguistics (Volume 1: Long Papers)}, pages 5755--5772, Dublin, Ireland, May 2022. Association for Computational Linguistics.

\bibitem{zhang-etal-2025-exploring}
Xinyu Zhang, Aibo Song, Jingyi Qiu, Jiahui Jin, Tianbo Zhang, and Xiaolin Fang.
\newblock Exploring multimodal relation extraction of hierarchical tabular data with multi-task learning.
\newblock In Wanxiang Che, Joyce Nabende, Ekaterina Shutova, and Mohammad~Taher Pilehvar, editors, {\em Proceedings of the 63rd Annual Meeting of the Association for Computational Linguistics (Volume 1: Long Papers)}, pages 26770--26781, Vienna, Austria, July 2025. Association for Computational Linguistics.

\bibitem{li2025mruiemultiperspectivereasoningreinforcement}
Zhongqiu Li, Shiquan Wang, Ruiyu Fang, Mengjiao Bao, Zhenhe Wu, Shuangyong Song, Yongxiang Li, and Zhongjiang He.
\newblock Mr-uie: Multi-perspective reasoning with reinforcement learning for universal information extraction, 2025.

\bibitem{fu-etal-2025-training}
Zichuan Fu, Xian Wu, Yejing Wang, Wanyu Wang, Shanshan Ye, Hongzhi Yin, Yi~Chang, Yefeng Zheng, and Xiangyu Zhao.
\newblock Training-free {LLM} merging for multi-task learning.
\newblock In Wanxiang Che, Joyce Nabende, Ekaterina Shutova, and Mohammad~Taher Pilehvar, editors, {\em Proceedings of the 63rd Annual Meeting of the Association for Computational Linguistics (Volume 1: Long Papers)}, pages 33111--33124, Vienna, Austria, July 2025. Association for Computational Linguistics.

\bibitem{Zheng2021MultimodalRE}
Changmeng Zheng, Junhao Feng, Ze~Fu, Yiru Cai, Qing Li, and Tao Wang.
\newblock Multimodal relation extraction with efficient graph alignment.
\newblock {\em Proceedings of the 29th ACM International Conference on Multimedia}, 2021.

\bibitem{luo-etal-2025-rear}
Jianwen Luo, Yu~Hong, Shuai Yang, and Jianmin Yao.
\newblock {REAR}: Reinforced reasoning optimization for event argument extraction with relation-aware support.
\newblock In Christos Christodoulopoulos, Tanmoy Chakraborty, Carolyn Rose, and Violet Peng, editors, {\em Findings of the Association for Computational Linguistics: EMNLP 2025}, pages 7957--7972, Suzhou, China, November 2025. Association for Computational Linguistics.

\bibitem{Wang2023InstructUIEMI}
Xiao Wang, Wei Zhou, Can Zu, Han Xia, Tianze Chen, Yuan Zhang, Rui Zheng, Junjie Ye, Qi~Zhang, Tao Gui, Jihua Kang, J.~Yang, Siyuan Li, and Chunsai Du.
\newblock Instructuie: Multi-task instruction tuning for unified information extraction.
\newblock {\em ArXiv}, abs/2304.08085, 2023.

\bibitem{doddington-etal-2004-automatic}
George Doddington, Alexis Mitchell, Mark Przybocki, Lance Ramshaw, Stephanie Strassel, and Ralph Weischedel.
\newblock The automatic content extraction ({ACE}) program {--} tasks, data, and evaluation.
\newblock In Maria~Teresa Lino, Maria~Francisca Xavier, F{\'a}tima Ferreira, Rute Costa, and Raquel Silva, editors, {\em Proceedings of the Fourth International Conference on Language Resources and Evaluation ({LREC}{'}04)}, Lisbon, Portugal, May 2004. European Language Resources Association (ELRA).

\bibitem{7780966}
Mark Yatskar, Luke Zettlemoyer, and Ali Farhadi.
\newblock Situation recognition: Visual semantic role labeling for image understanding.
\newblock In {\em 2016 IEEE Conference on Computer Vision and Pattern Recognition (CVPR)}, pages 5534--5542, 2016.

\bibitem{10.1145/3474085.3476968}
Changmeng Zheng, Junhao Feng, Ze~Fu, Yi~Cai, Qing Li, and Tao Wang.
\newblock Multimodal relation extraction with efficient graph alignment.
\newblock In {\em Proceedings of the 29th ACM International Conference on Multimedia}, MM '21, page 5298–5306, New York, NY, USA, 2021. Association for Computing Machinery.

\bibitem{10.1007/978-3-030-58548-8_19}
Sarah Pratt, Mark Yatskar, Luca Weihs, Ali Farhadi, and Aniruddha Kembhavi.
\newblock Grounded situation recognition.
\newblock In Andrea Vedaldi, Horst Bischof, Thomas Brox, and Jan-Michael Frahm, editors, {\em Computer Vision -- ECCV 2020}, pages 314--332, Cham, 2020. Springer International Publishing.

\bibitem{wang2024qwen2vlenhancingvisionlanguagemodels}
Peng Wang, Shuai Bai, Sinan Tan, Shijie Wang, Zhihao Fan, Jinze Bai, Keqin Chen, Xuejing Liu, Jialin Wang, Wenbin Ge, Yang Fan, Kai Dang, Mengfei Du, Xuancheng Ren, Rui Men, Dayiheng Liu, Chang Zhou, Jingren Zhou, and Junyang Lin.
\newblock Qwen2-vl: Enhancing vision-language model's perception of the world at any resolution, 2024.

\bibitem{bai2025qwen3vltechnicalreport}
Shuai Bai, Yuxuan Cai, Ruizhe Chen, Keqin Chen, Xionghui Chen, Zesen Cheng, Lianghao Deng, Wei Ding, Chang Gao, Chunjiang Ge, Wenbin Ge, Zhifang Guo, Qidong Huang, Jie Huang, Fei Huang, Binyuan Hui, Shutong Jiang, Zhaohai Li, Mingsheng Li, Mei Li, Kaixin Li, Zicheng Lin, Junyang Lin, Xuejing Liu, Jiawei Liu, Chenglong Liu, Yang Liu, Dayiheng Liu, Shixuan Liu, Dunjie Lu, Ruilin Luo, Chenxu Lv, Rui Men, Lingchen Meng, Xuancheng Ren, Xingzhang Ren, Sibo Song, Yuchong Sun, Jun Tang, Jianhong Tu, Jianqiang Wan, Peng Wang, Pengfei Wang, Qiuyue Wang, Yuxuan Wang, Tianbao Xie, Yiheng Xu, Haiyang Xu, Jin Xu, Zhibo Yang, Mingkun Yang, Jianxin Yang, An~Yang, Bowen Yu, Fei Zhang, Hang Zhang, Xi~Zhang, Bo~Zheng, Humen Zhong, Jingren Zhou, Fan Zhou, Jing Zhou, Yuanzhi Zhu, and Ke~Zhu.
\newblock Qwen3-vl technical report, 2025.

\bibitem{wang2025internvl35advancingopensourcemultimodal}
Weiyun Wang, Zhangwei Gao, Lixin Gu, Hengjun Pu, Long Cui, Xingguang Wei, Zhaoyang Liu, Linglin Jing, Shenglong Ye, Jie Shao, Zhaokai Wang, Zhe Chen, Hongjie Zhang, Ganlin Yang, Haomin Wang, Qi~Wei, Jinhui Yin, Wenhao Li, Erfei Cui, Guanzhou Chen, Zichen Ding, Changyao Tian, Zhenyu Wu, Jingjing Xie, Zehao Li, Bowen Yang, Yuchen Duan, Xuehui Wang, Zhi Hou, Haoran Hao, Tianyi Zhang, Songze Li, Xiangyu Zhao, Haodong Duan, Nianchen Deng, Bin Fu, Yinan He, Yi~Wang, Conghui He, Botian Shi, Junjun He, Yingtong Xiong, Han Lv, Lijun Wu, Wenqi Shao, Kaipeng Zhang, Huipeng Deng, Biqing Qi, Jiaye Ge, Qipeng Guo, Wenwei Zhang, Songyang Zhang, Maosong Cao, Junyao Lin, Kexian Tang, Jianfei Gao, Haian Huang, Yuzhe Gu, Chengqi Lyu, Huanze Tang, Rui Wang, Haijun Lv, Wanli Ouyang, Limin Wang, Min Dou, Xizhou Zhu, Tong Lu, Dahua Lin, Jifeng Dai, Weijie Su, Bowen Zhou, Kai Chen, Yu~Qiao, Wenhai Wang, and Gen Luo.
\newblock Internvl3.5: Advancing open-source multimodal models in versatility, reasoning, and efficiency, 2025.

\bibitem{radford2021learningtransferablevisualmodels}
Alec Radford, Jong~Wook Kim, Chris Hallacy, Aditya Ramesh, Gabriel Goh, Sandhini Agarwal, Girish Sastry, Amanda Askell, Pamela Mishkin, Jack Clark, Gretchen Krueger, and Ilya Sutskever.
\newblock Learning transferable visual models from natural language supervision, 2021.

\bibitem{devlin-etal-2019-bert}
Jacob Devlin, Ming-Wei Chang, Kenton Lee, and Kristina Toutanova.
\newblock {BERT}: Pre-training of deep bidirectional transformers for language understanding.
\newblock In Jill Burstein, Christy Doran, and Thamar Solorio, editors, {\em Proceedings of the 2019 Conference of the North {A}merican Chapter of the Association for Computational Linguistics: Human Language Technologies, Volume 1 (Long and Short Papers)}, pages 4171--4186, Minneapolis, Minnesota, June 2019. Association for Computational Linguistics.

\bibitem{DosovitskiyB0WZ21}
Alexey Dosovitskiy, Lucas Beyer, Alexander Kolesnikov, Dirk Weissenborn, Xiaohua Zhai, Thomas Unterthiner, Mostafa Dehghani, Matthias Minderer, Georg Heigold, Sylvain Gelly, Jakob Uszkoreit, and Neil Houlsby.
\newblock An image is worth 16x16 words: Transformers for image recognition at scale.
\newblock In {\em 9th International Conference on Learning Representations, {ICLR} 2021, Virtual Event, Austria, May 3-7, 2021}. OpenReview.net, 2021.

\bibitem{10533619}
Rejin Varghese and Sambath M.
\newblock Yolov8: A novel object detection algorithm with enhanced performance and robustness.
\newblock In {\em 2024 International Conference on Advances in Data Engineering and Intelligent Computing Systems (ADICS)}, pages 1--6, 2024.

\bibitem{chung2022scalinginstructionfinetunedlanguagemodels}
Hyung~Won Chung, Le~Hou, Shayne Longpre, Barret Zoph, Yi~Tay, William Fedus, Yunxuan Li, Xuezhi Wang, Mostafa Dehghani, Siddhartha Brahma, Albert Webson, Shixiang~Shane Gu, Zhuyun Dai, Mirac Suzgun, Xinyun Chen, Aakanksha Chowdhery, Alex Castro-Ros, Marie Pellat, Kevin Robinson, Dasha Valter, Sharan Narang, Gaurav Mishra, Adams Yu, Vincent Zhao, Yanping Huang, Andrew Dai, Hongkun Yu, Slav Petrov, Ed~H. Chi, Jeff Dean, Jacob Devlin, Adam Roberts, Denny Zhou, Quoc~V. Le, and Jason Wei.
\newblock Scaling instruction-finetuned language models, 2022.

\bibitem{JMLR:v21:20-074}
Colin Raffel, Noam Shazeer, Adam Roberts, Katherine Lee, Sharan Narang, Michael Matena, Yanqi Zhou, Wei Li, and Peter~J. Liu.
\newblock Exploring the limits of transfer learning with a unified text-to-text transformer.
\newblock {\em Journal of Machine Learning Research}, 21(140):1--67, 2020.

\bibitem{Liu_2024_CVPR}
Haotian Liu, Chunyuan Li, Yuheng Li, and Yong~Jae Lee.
\newblock Improved baselines with visual instruction tuning.
\newblock In {\em Proceedings of the IEEE/CVF Conference on Computer Vision and Pattern Recognition (CVPR)}, pages 26296--26306, June 2024.

\end{thebibliography}

\end{document}